\documentclass[journal]{IEEEtran}
\usepackage{amsmath,amsfonts,amssymb}
\usepackage{algorithmic}
\usepackage{algorithm}
\usepackage{array}
\usepackage[caption=false,font=normalsize,labelfont=sf,textfont=sf]{subfig}
\usepackage{textcomp}
\usepackage{stfloats}
\usepackage{url}
\usepackage{verbatim}
\usepackage{graphicx}
\usepackage{cite}
\usepackage{multirow}
\usepackage[table]{xcolor}

\usepackage{color}
\usepackage{booktabs}
\usepackage{makecell}

\usepackage[colorlinks,bookmarksopen,bookmarksnumbered,citecolor=green, linkcolor=red, urlcolor=magenta]{hyperref}

\begin{document}

\title{Evaluating Image Editing with LLMs: A Comprehensive Benchmark \\ and Intermediate-Layer Probing Approach}

\author{
Shiqi Gao, Zitong Xu, Kang Fu, Huiyu Duan, Xiongkuo Min, Jia Wang\IEEEauthorrefmark{2}%
\thanks{\IEEEauthorrefmark{2} Co-corresponding authors.}%
\thanks{Shiqi Gao, Zitong Xu, Kang Fu, Huiyu Duan, Xiongkuo Min and Jia Wang are with the Institute of Image Communication and Network Engineering, Shanghai Jiao Tong University, Shanghai 200240, China (e-mail: gaoshiqi@sjtu.edu.cn; xuzitong@sjtu.edu.cn; fuk20-20@sjtu.edu.cn; huiyuduan@sjtu.edu.cn; minxiongkuo@sjtu.edu.cn; jiawang@sjtu.edu.cn).}%
}



\maketitle

\begin{abstract}
Evaluating text-guided image editing (TIE) methods remains a challenging problem, as reliable assessment should simultaneously consider perceptual quality, alignment with textual instructions, and preservation of original image content. Despite rapid progress in TIE models, existing evaluation benchmarks remain limited in scale and often show weak correlation with human perceptual judgments. In this work, we introduce TIEdit, a benchmark for systematic evaluation of text-guided image editing methods. TIEdit consists of 512 source images paired with editing prompts across eight representative editing tasks, producing 5,120 edited images generated by ten state-of-the-art TIE models. To obtain reliable subjective ratings, 20 experts are recruited to produce 307,200 raw subjective ratings, which accumulates into 15,360 mean opinion scores (MOSs) across three evaluation dimensions: perceptual quality, editing alignment, and content preservation. Beyond the benchmark itself, we further propose EditProbe, an LLM-based evaluator that estimates editing quality via intermediate-layer probing of hidden representations. Instead of relying solely on final model outputs, EditProbe extracts informative representations from intermediate layers of multimodal large language models to better capture semantic and perceptual relationships between source images, editing instructions, and edited results. Experimental results demonstrate that widely used automatic evaluation metrics show limited correlation with human judgments on editing tasks, while EditProbe achieves substantially stronger alignment with human perception. Together, TIEdit and EditProbe provide a foundation for more reliable and perceptually aligned evaluation of text-guided image editing methods.
\end{abstract}




\section{Introduction}

Text-based image editing (TIE) has recently emerged as a powerful paradigm for controlling image manipulation through natural language instructions \cite{stable_diffusion,instructpix2pix,text2live}. With the rapid progress of diffusion-based generative models, modern TIE methods can perform diverse editing operations, including object removal, attribute modification, scene reconfiguration, and style transformation. These advances significantly broaden the applicability of image editing technologies in creative design, content generation, and human--computer interaction. 

Despite these advances, evaluating the quality and correctness of text-based edits remains challenging. Unlike conventional image generation tasks, TIE requires models to perform targeted modifications while preserving the overall structure and irrelevant content of the original image. Consequently, although numerous methods have been developed to evaluate AI-generated images \cite{wang2025lmm4lmm,liu2025f,wang2026quality,wang2023aigciqa2023}, for TIE evaluation, we need evaluation metrics that reliable evaluation must jointly consider perceptual quality, alignment with the editing instruction, and preservation of unmodified image regions.

Existing evaluation approaches only partially address these requirements for TIE assessment. Several benchmarks have been proposed for text-guided editing evaluation \cite{tedbench,editval,i2ebench,emuedit,lmm4edit,wang2025i2i}, yet many of them cover limited editing scenarios or rely heavily on automatic scoring. Meanwhile, widely used automatic metrics, including traditional image quality assessment measures such as SSIM \cite{wang2004ssim} and generative evaluation metrics such as FID \cite{heusel2017gans}, primarily capture visual similarity or global realism rather than editing correctness. Even recent vision--language metrics, such as CLIPScore and ImageReward \cite{clipscore,database/align:ImageReward}, cannot explicitly evaluate whether the intended modification described by the instruction has been correctly applied to the source image. As a result, existing automatic metrics often exhibit weak correlation with human perceptual judgments.

\begin{figure*}[t]
    \centering
    \includegraphics[width=0.99\linewidth]{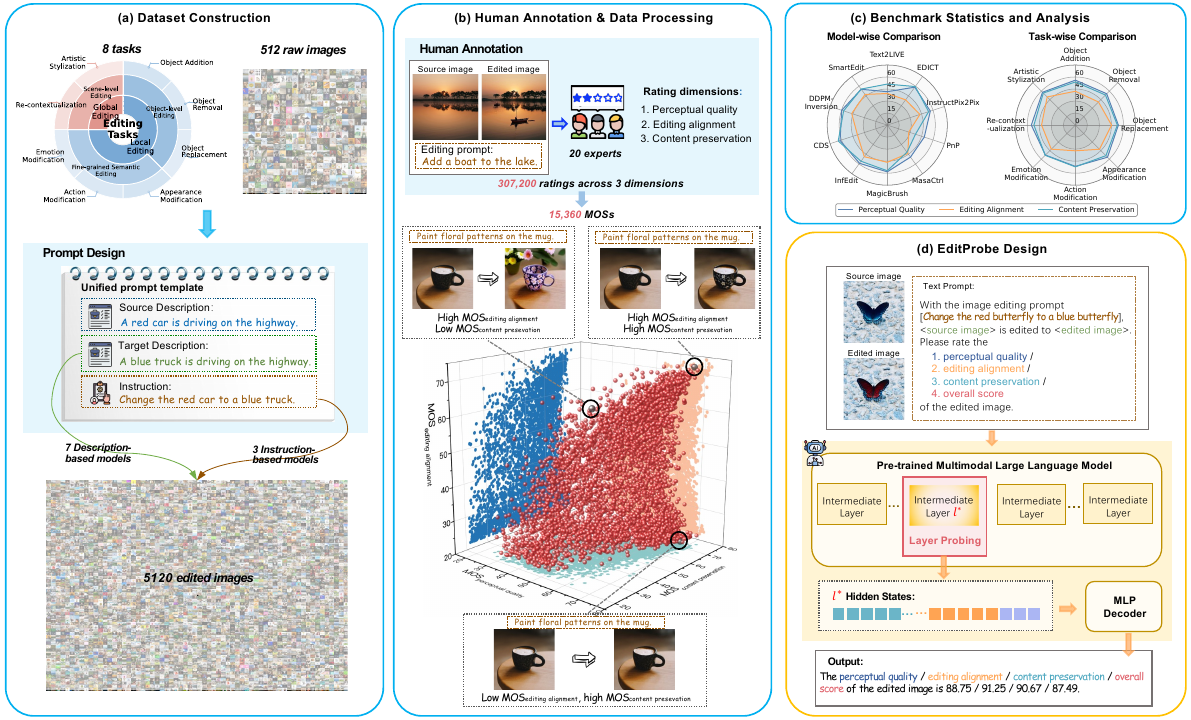}
    \vspace{-2pt}
    \caption{
Overview of the proposed TIEdit benchmark and the EditProbe evaluation method. 
(a) We collect 512 source images and design editing prompts covering eight representative editing tasks. 
Ten text-guided image editing models are applied to generate 5,120 edited images. 
(b) 307,200 human ratings are collected from 20 expert annotators across three evaluation dimensions and processed to obtain 15,360 MOSs. 
(c) We analyze the benchmark from three perspectives: task-wise, model-wise, and dimension-wise.
(d) We propose EditProbe to predict editing quality through intermediate-layer probing of a multimodal LLM.
}
    \vspace{-15pt}
    \label{fig:1}
\end{figure*}

To address these limitations, we introduce \textbf{TIEdit}, a benchmark for systematic evaluation of text-based image editing. TIEdit contains 512 source images paired with editing prompts across eight representative editing tasks, producing 5,120 edited images generated by ten state-of-the-art TIE models. A large-scale subjective study involving 20 expert annotators was conducted over five months, resulting in 15,360 mean opinion scores derived from 307,200 raw subjective ratings across three evaluation dimensions, \textit{i.e.}, perceptual quality, editing alignment, and content preservation. The benchmark further incorporates spatially enriched editing prompts that explicitly describe spatial relationships and geometric transformations, enabling more rigorous evaluation of spatial editing accuracy.

Beyond constructing the benchmark, we propose \textbf{EditProbe}, a LLM-based evaluation framework for predicting editing quality. Instead of relying solely on the final outputs of multimodal large language models, EditProbe probes intermediate transformer layers to extract representations that capture the relationships among the source image, editing instruction, and edited result. Based on these representations, lightweight regression heads are trained to predict multi-dimensional editing quality scores aligned with human judgments.
Extensive experiments on TIEdit reveal a substantial gap between widely used automatic metrics and human perceptual evaluations. In contrast, EditProbe achieves significantly stronger correlation with human judgments, demonstrating the effectiveness of intermediate-layer probing for editing quality assessment. An overview of the proposed TIEdit benchmark and the EditProbe evaluation framework is illustrated in Fig.~\ref{fig:1}.

The main contributions of this paper are summarized as follows:

\begin{itemize}

\item We introduce TIEdit, a benchmark for text-based image editing evaluation containing 5,120 edited images generated by ten editing models and annotated with 15,360 MOSs obtained from a large-scale human study.

\item Using TIEdit, we analyze the performance of existing automatic evaluation metrics and reveal their limitations in capturing human perceptual judgments for editing tasks.

\item We propose EditProbe, an LLM-based evaluation framework that predicts editing quality via intermediate-layer probing, achieving improved alignment with human perceptual evaluations.

\end{itemize}

\section{Related Work}
\label{sec:related}

\subsection{Text-guided Image Editing}

Text-guided image editing (TIE) has advanced rapidly with the development of large-scale generative models and multimodal representation learning. Recent progress in generative modeling, particularly diffusion-based frameworks such as Stable Diffusion~\cite{stable_diffusion} and FLUX~\cite{flux}, has substantially improved the controllability and realism of text-conditioned visual synthesis. Building upon these advances, TIE methods have emerged as an important research direction for modifying existing images according to either textual descriptions or editing instructions.

Existing TIE methods can generally be categorized into description-based and instruction-based approaches. Description-based methods formulate editing as a transformation from a source description to a target description, where the image is modified to better match the target prompt. A representative line of research relies on optimization- or inversion-based mechanisms to project the source image into a latent space and then regenerate the edited result under the target prompt, such as Text2LIVE~\cite{text2live}, EDICT~\cite{edict}, DDPM inversion-based editing~\cite{ddpm_edit}, MasaCtrl~\cite{masactrl}, PnP inversion~\cite{pnp_inversion}, Contrastive Denoising Score (CDS)~\cite{cds}, InfEdit~\cite{infedit}, ReNoise~\cite{renoise}, RFSE~\cite{rfse}, and FlowEdit~\cite{flowedit}. In contrast, instruction-based methods directly follow natural language editing commands, making them more flexible and more closely aligned with practical image editing scenarios. InstructPix2Pix~\cite{instructpix2pix} is a representative example, and subsequent methods such as MagicBrush~\cite{magicbrush}, Any2Pix~\cite{any2pix}, ZONE~\cite{zone}, HQ-Edit~\cite{hqedit}, and ACE++~\cite{acepp} further improve editing fidelity, controllability, and instruction understanding.

The development of large vision-language models has also strengthened the alignment between textual instructions and visual representations. Models such as CLIP~\cite{radford2021learning}, BLIP~\cite{li2022blip}, BLIP-2~\cite{li2023blip2}, InstructBLIP~\cite{instructblip}, LLaVA~\cite{liu2023llava}, InternVL~\cite{internvl}, and Qwen2-VL~\cite{qwen2vl} have provided stronger multimodal representations for both editing and evaluation. Nevertheless, TIE remains challenging because successful editing requires not only following the instruction, but also preserving perceptual quality and maintaining consistency with the original image, particularly for spatially explicit instructions and complex scene-level modifications.

\subsection{Text-based Image Editing Assessment}

Assessing text-guided image editing is substantially more difficult than evaluating conventional image generation. Existing assessment methods can be broadly divided into three categories: dedicated TIE benchmarks, traditional and multimodal evaluation metrics, and recent LLM-based evaluators.

A growing number of benchmarks have been proposed for TIE evaluation, but most remain limited in scale, task coverage, or annotation depth. TedBench~\cite{tedbench}, one of the earliest TIE benchmarks, contains only a very small set of source images and edited outputs. EditVal~\cite{editval} extends the task scope but is constrained by the quality of generated images. EditEval~\cite{survey_diffusion_editing} summarizes broader TIE settings but does not provide perceptual ground-truth labels aligned with human judgments. I2EBench~\cite{i2ebench} introduces a comprehensive benchmark for instruction-based image editing, while Emu Edit~\cite{emuedit} provides both instruction and description prompts across multiple editing scenarios. More recently, LMM4Edit~\cite{lmm4edit} proposes a multimodal benchmark together with an LMM-based evaluator that predicts perceptual quality, editing alignment, and attribute preservation. These efforts highlight the importance of benchmark-driven and human-aligned evaluation, but there remains room for improvement in benchmark design, annotation scale, and automatic evaluation methodology.

In parallel, a large body of image quality assessment (IQA) methods has been adopted as automatic baselines. Full-reference metrics, such as SSIM~\cite{wang2004ssim}, MS-SSIM~\cite{wang2003msssim}, FSIM~\cite{zhang2011fsim}, and IFC~\cite{sheikh2005ifc}, measure fidelity relative to a reference image, while no-reference methods, such as BRISQUE~\cite{mittal2012making}, NIQE~\cite{mittal2012niqe}, CNNIQA~\cite{kang2014convolutional}, HyperIQA~\cite{su2020blindly}, DBCNN~\cite{quality:DBCNN}, MANIQA~\cite{yang2022maniqa}, NIMA~\cite{talebi2018nima}, and TOPIQ~\cite{topiq}, estimate perceptual quality without a pristine reference. However, these methods were not designed for TIE and therefore cannot explicitly assess whether the intended edit has been correctly applied or whether unedited content has been properly preserved.

To better capture image-text consistency, multimodal metrics such as CLIPScore~\cite{clipscore}, ImageReward~\cite{database/align:ImageReward}, and Pick-a-Pic / PickScore~\cite{database/align:PickAPic} have been introduced. Although effective for text-to-image generation assessment, these methods are less suitable for editing tasks, where evaluation must jointly consider the source image, the edited result, and the editing instruction. Recently, LLM-based assessment has attracted increasing attention, motivated by the strong reasoning and multimodal understanding capabilities of large language models. LMM4Edit \cite{lmm4edit} shows that large multimodal models \cite{duan2025finevq,liu2025moa} can serve as effective evaluators for image editing, yet the potential of exploiting intermediate hidden representations for quality prediction remains underexplored.

Overall, current methodologies remain insufficient for jointly capturing perceptual quality, editing alignment, and content preservation in text-guided image editing. These limitations motivate the development of more comprehensive benchmarks and more reliable automatic evaluators, which form the basis of the proposed TIEdit benchmark and EditProbe framework.

\begin{table*}[!t]
\centering
\caption{Summary of the eight editing tasks defined in the TIEdit benchmark.}
\label{tab:task}
\vspace{-4pt}
\begin{tabular}{l l c}
\hline
\textbf{Task} & \textbf{Description} & \textbf{Scope} \\
\hline
Object Addition & Insert a new object into the scene. & Local \\
Object Removal & Remove an object and restore the background. & Local \\
Object Replacement & Replace an object with another one. & Local \\
Appearance Modification & Change object color, material, or texture. & Local \\
Action Modification & Modify the pose or action of a subject. & Local \\
Emotion Modification & Change facial expression or affective cues. & Local \\
Re-contextualization & Alter the global environment or scene context. & Global \\
Artistic Stylization & Transform the image into an artistic style. & Global \\
\hline
\end{tabular}
\vspace{-10pt}
\end{table*}

\section{TIEdit Benchmark}
\label{sec:bench}
This section presents the construction of the proposed TIEdit benchmark for evaluating text-guided image editing methods. 
Recent studies suggest that reliable evaluation of text-guided image editing requires benchmarks that cover diverse editing scenarios and include large-scale human perceptual annotations \cite{editval,i2ebench,emuedit}. 
As illustrated in Fig.~\ref{fig:1}(a)(b), the TIEdit benchmark is constructed through a multi-stage pipeline including editing task design, benchmark construction, and large-scale subjective evaluation. 
In total, TIEdit contains 512 source images paired with editing prompts spanning eight representative editing tasks. 
Ten state-of-the-art text-guided image editing models are applied to generate 5,120 edited images. 
Through a five-month subjective evaluation campaign involving 20 expert annotators, we obtain MOSs across three evaluation dimensions: perceptual quality, editing alignment, and content preservation.

\subsection{Editing Task Design}

A comprehensive benchmark for text-guided image editing should cover diverse editing scenarios encountered in practical applications. 
Previous works emphasize that evaluating editing models across multiple semantic transformation types is necessary to fully assess editing capability \cite{editval,i2ebench}. 
These tasks cover both local object-level edits and global scene-level transformations. 
Object-level operations such as addition, removal, and replacement evaluate whether models can manipulate visual entities while maintaining spatial coherence with surrounding content. 
Appearance, action, and emotion modification assess the ability of editing models to perform fine-grained semantic transformations without introducing structural artifacts. 
In contrast, re-contextualization and artistic stylization involve more global modifications that alter the scene environment or artistic style, requiring stronger semantic understanding and visual consistency.
Table~\ref{tab:task} summarizes the task definitions used in the TIEdit benchmark.

\subsection{Benchmark Construction}

After defining the editing tasks, we construct the benchmark by collecting source images, generating editing prompts, and producing edited images using multiple text-guided image editing models.

\subsubsection{Image Sources}

We collect 512 high-quality source images from the public photography platform Unsplash\cite{unsplash}, which provides diverse real-world scenes including people, animals, urban environments, indoor scenes, and natural landscapes. 
These images are selected to ensure broad visual diversity and realistic content suitable for editing evaluation. 
To maintain compatibility with most diffusion-based editing frameworks \cite{rombach2022stable}, all images are cropped into square formats.

\subsubsection{Prompt Generation}
For each source image, we construct an editing prompt that describes the intended transformation. 
Each prompt contains three textual components: a source description describing the original image content, a target description specifying the desired edited result, and an editing instruction that explicitly defines the modification operation. 
This design enables the benchmark to support both description-based and instruction-based editing models.

To improve linguistic consistency and prompt diversity, GPT-4~\cite{openai2023gpt4} is used to generate candidate prompts based on a structured template, following recent practices in LLM-assisted dataset construction \cite{instructpix2pix,magicbrush}. 
The generated prompts are subsequently reviewed and refined to ensure correctness and clarity. 
Approximately 10\% of the prompts intentionally include more complex editing requirements involving spatial relationships or compositional reasoning. 
Previous studies have shown that spatial reasoning and compositional transformations remain challenging for text-guided image generation and editing models \cite{gokhale2022benchmark,liu2024reasoning}. 
Including such prompts enables the benchmark to evaluate whether editing models can correctly interpret spatial constraints and perform compositional edits.

\subsubsection{Text-guided Image Editing Models}

To construct a representative benchmark, we include ten recent text-guided image editing models that cover different editing paradigms. 
Specifically, the selected models include Text2LIVE~\cite{text2live}, EDICT~\cite{edict}, InstructPix2Pix~\cite{instructpix2pix}, Plug-and-Play Diffusion (PnP)~\cite{pnpdiffusion}, MasaCtrl~\cite{masactrl}, MagicBrush~\cite{magicbrush}, InfEdit~\cite{infedit}, CDS~\cite{cds}, DDPM-Inversion~\cite{ddpminversion}, and SmartEdit~\cite{smartedit}. 
These models represent two mainstream paradigms in text-guided image editing: diffusion-based editing and instruction-following editing.

Among these models, Text2LIVE~\cite{text2live}, EDICT~\cite{edict}, PnP~\cite{pnpdiffusion}, MasaCtrl~\cite{masactrl}, InfEdit~\cite{infedit}, CDS~\cite{cds}, and DDPM-Inversion~\cite{ddpminversion} follow a description-based editing paradigm where the editing process is guided by textual descriptions of the desired output image. 
In contrast, InstructPix2Pix~\cite{instructpix2pix}, MagicBrush~\cite{magicbrush}, and SmartEdit~\cite{smartedit} adopt an instruction-based editing paradigm in which the model directly follows natural language editing instructions.

To ensure that the benchmark can be applied to all selected models, we design editing prompts in a unified and compatible format. 
For description-based models, the target description describing the edited image is used as input. 
For instruction-based models, the corresponding editing instruction is used instead. 
This unified prompt design allows the same benchmark samples to be seamlessly adapted to different editing paradigms while maintaining consistent editing semantics across models.
Using the collected source images and editing prompts, each model generates edited outputs for all benchmark samples, producing a total of 5,120 edited images.

\subsection{Subjective Experiment}

{\bfseries Experimental environment and apparatus.}
Following the recommendations of the ITU-R BT.500 standard for subjective visual quality assessment \cite{itu500}, the subjective experiment was conducted in a controlled environment to minimize the influence of ambient lighting and display variations. The evaluation was performed using a 27-inch LCD monitor with a resolution of $3840 \times 2160$. The display color temperature was set to 6500K and the brightness level was fixed at 50 \cite{zhou2019no}.
Participants were seated at a viewing distance of approximately 70\,cm from the screen, which allows comfortable observation and comparison of images. 

{\bfseries Experimental procedure.}
We conduct a subjective quality
assessment experiment\cite{itu500,p910}. 
participants are presented with a source image, an edited image, and the corresponding editing prompt for each evaluation.
They are asked to rate the edited images using a five-point scale across three dimensions: perceptual quality, editing alignment, and content preservation. 
Specifically, perceptual quality measures the overall visual quality and naturalness of the edited image, editing alignment evaluates whether the generated edit correctly follows the given instruction, and content preservation assesses the extent to which the original image content that should remain unchanged is preserved.

Before the evaluation began, 20 training image pairs with different editing qualities were shown to familiarize participants with the rating criteria. During the experiment, participants were encouraged to take short breaks every 20 minutes to avoid visual fatigue. The entire annotation campaign lasted approximately five months.

{\bfseries Subjective Data Processing.}
According to the recommendation in \cite{itu500}, at least 15 subjects are required to obtain reliable subjective evaluation results. In this work, 20 expert annotators with experience in photography and visual quality evaluation participated in the experiment.

Outlier detection and subject rejection were conducted following the guidelines in \cite{itu500,duan2022confusing}. Specifically, we first examine the kurtosis of the 307,200 raw ratings to determine whether the score distribution follows a Gaussian or non-Gaussian distribution. For Gaussian cases, ratings that deviate by more than two standard deviations from the image mean are treated as outliers. For non-Gaussian distributions, the $\sqrt{20}$ criterion is applied as described in \cite{duan2022confusing}. Subjects whose rating deviations exceeded 5\% were rejected. As a result, one subject was removed from the dataset, and approximately 1.89\% of the ratings were identified as outliers and discarded.

Finally, the mean opinion scores (MOSs) are obtained after subject normalization using the following equations:

\begin{equation}
z_{ij}=\frac{m_{ij}-\mu_i}{\sigma_i}, \quad
z_{ij}'=\frac{100(z_{ij}+3)}{6},
\end{equation}
\begin{equation}
MOS_j=\frac{1}{N}\sum_{i=1}^{N} z_{ij}',
\end{equation}

where $m_{ij}$ denotes the raw rating assigned by the $i$-th subject to the $j$-th image, $\mu_i$ and $\sigma_i$ denote the mean and standard deviation of the ratings from subject $i$, and $N$ is the number of valid subjects.

For convenience, the MOSs corresponding to the three evaluation dimensions are denoted as $MOS_q$, $MOS_e$, and $MOS_p$, representing perceptual quality, editing alignment, and content preservation, respectively. In addition, an overall score $MOS_o$ is computed as the average of the three dimensions, i.e.,
\[
MOS_o = \frac{MOS_q + MOS_e + MOS_p}{3},
\]
which provides a holistic evaluation of editing quality.

\begin{figure}[!t]
    \centering
    \includegraphics[width=0.91\linewidth]{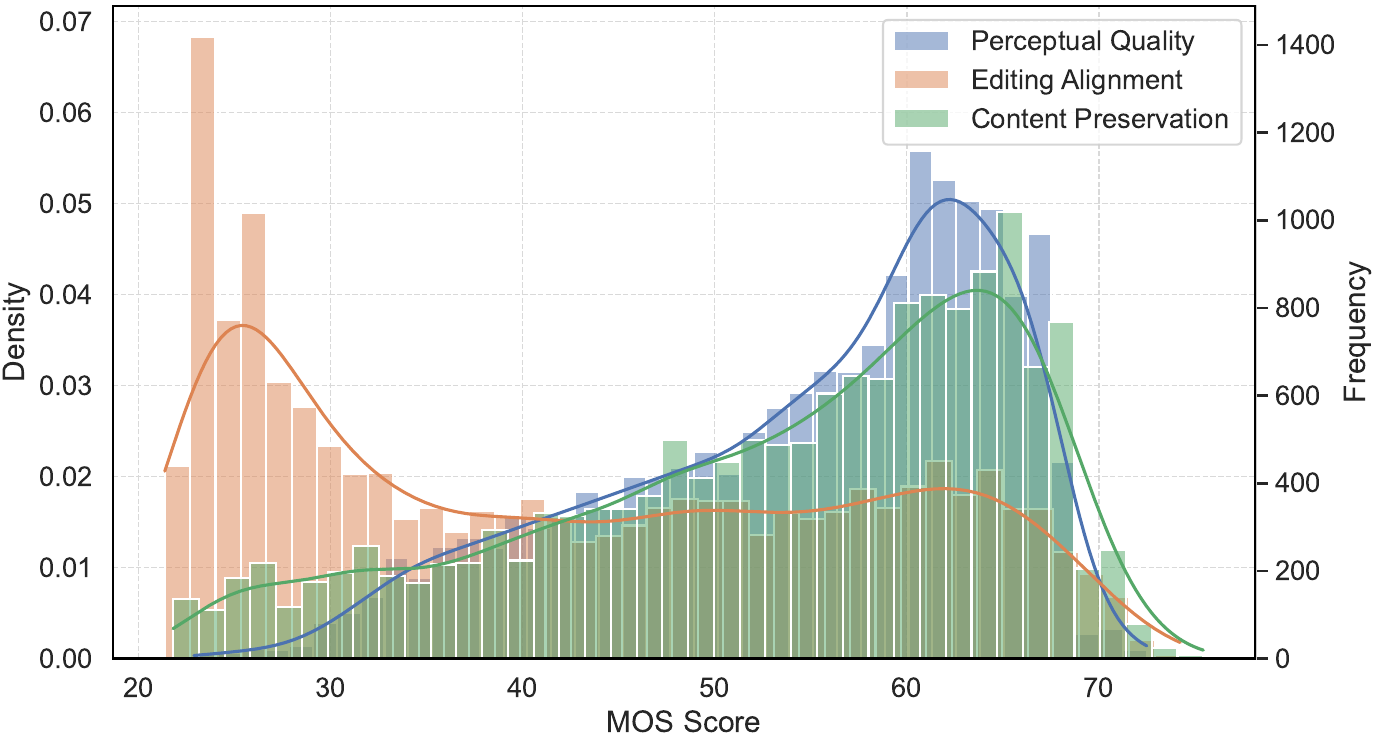}
    \vspace{-0pt}
    \caption{
Distribution of MOSs across the three evaluation dimensions in the TIEdit benchmark.
The histogram and kernel density curves illustrate the global score distributions for perceptual quality, editing alignment, and content preservation.
Perceptual quality and content preservation generally concentrate at higher score ranges, while editing alignment exhibits a broader distribution, indicating greater difficulty in accurately following editing instructions.
}
    \vspace{-0pt}
    \label{fig:mos_distribution}
\end{figure}

\begin{figure*}[t]
    \centering
    \includegraphics[width=0.86\linewidth]{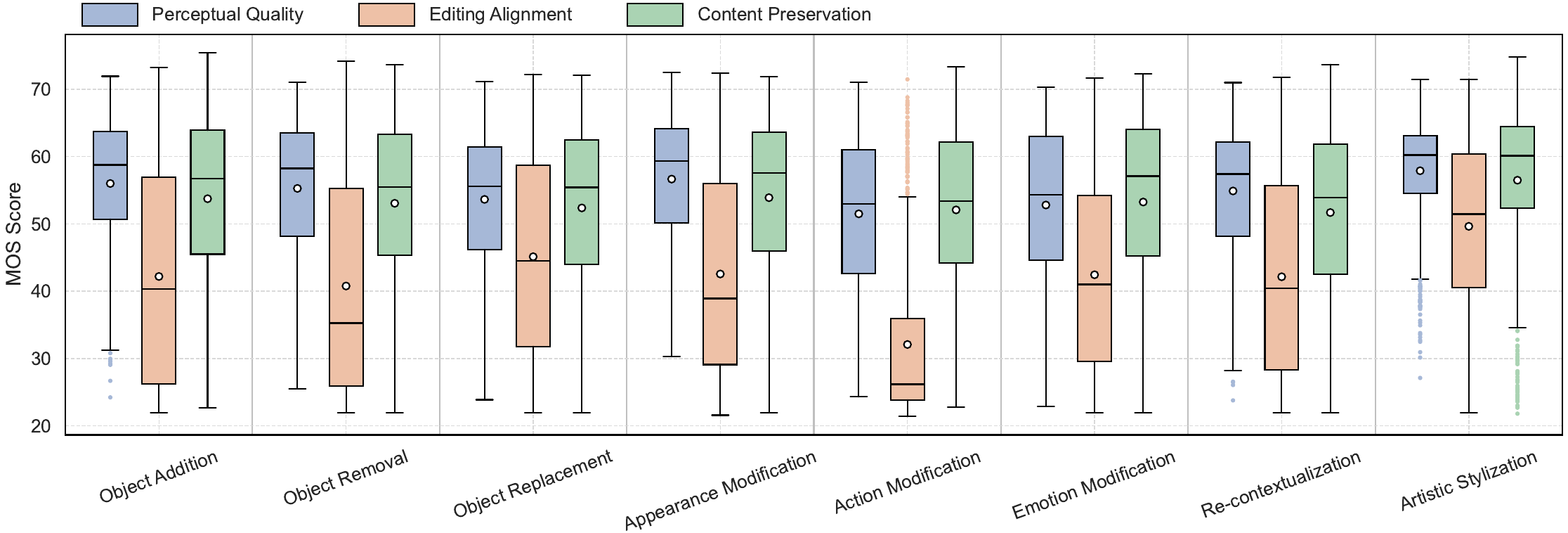}
    \vspace{-0pt}
    \caption{
MOS distributions across different editing tasks in the TIEdit benchmark.
Each boxplot summarizes the subjective scores for perceptual quality, editing alignment, and content preservation within a specific editing task.
Tasks involving complex semantic transformations, such as action modification and emotion modification, tend to exhibit lower editing alignment scores, indicating increased editing difficulty.
}
    \vspace{-0pt}
    \label{fig:taskwise}
\end{figure*}

\subsection{Benchmark Statistics and Analysis}

To better understand the characteristics of the TIEdit benchmark and the behavior of existing text-guided image editing models, we analyze the subjective evaluation results from multiple perspectives, including the global MOS distributions, task-wise difficulty, model-wise performance, and dimension-wise characteristics.

\textbf{Global MOS Distribution.}
We first analyze the global distribution of MOSs across the three evaluation dimensions. 
As shown in Fig.~\ref{fig:mos_distribution}, the MOS values span a wide range across perceptual quality, editing alignment, and content preservation, indicating diverse editing outcomes in the benchmark. 
In general, perceptual quality and content preservation tend to concentrate at relatively higher score ranges, suggesting that many edited results maintain acceptable visual realism and structural consistency. 
In contrast, editing alignment exhibits a broader distribution with noticeably lower average scores. 
This observation indicates that accurately following editing instructions remains a major challenge for current text-guided image editing models, which is consistent with findings reported in previous editing benchmarks \cite{editval,i2ebench,magicbrush}.

\textbf{Task-wise Analysis.}
To investigate how editing difficulty varies across editing scenarios, we analyze MOS distributions for each editing task. 
Fig.~\ref{fig:taskwise} presents the MOS statistics across the eight editing tasks. 
Tasks involving relatively localized modifications, such as object addition, object removal, and object replacement, generally exhibit more stable score distributions. 
In contrast, tasks requiring more complex semantic transformations, including action modification, emotion modification, and re-contextualization, show larger performance variance across models. 
In particular, re-contextualization and emotion modification tend to produce lower editing alignment scores, suggesting that edits requiring deeper semantic reasoning and contextual understanding remain more challenging for current models \cite{gokhale2022benchmark,liu2024reasoning}.

\begin{figure*}[!t]
    \centering
    \vspace{-3pt}
    \includegraphics[width=0.93\linewidth]{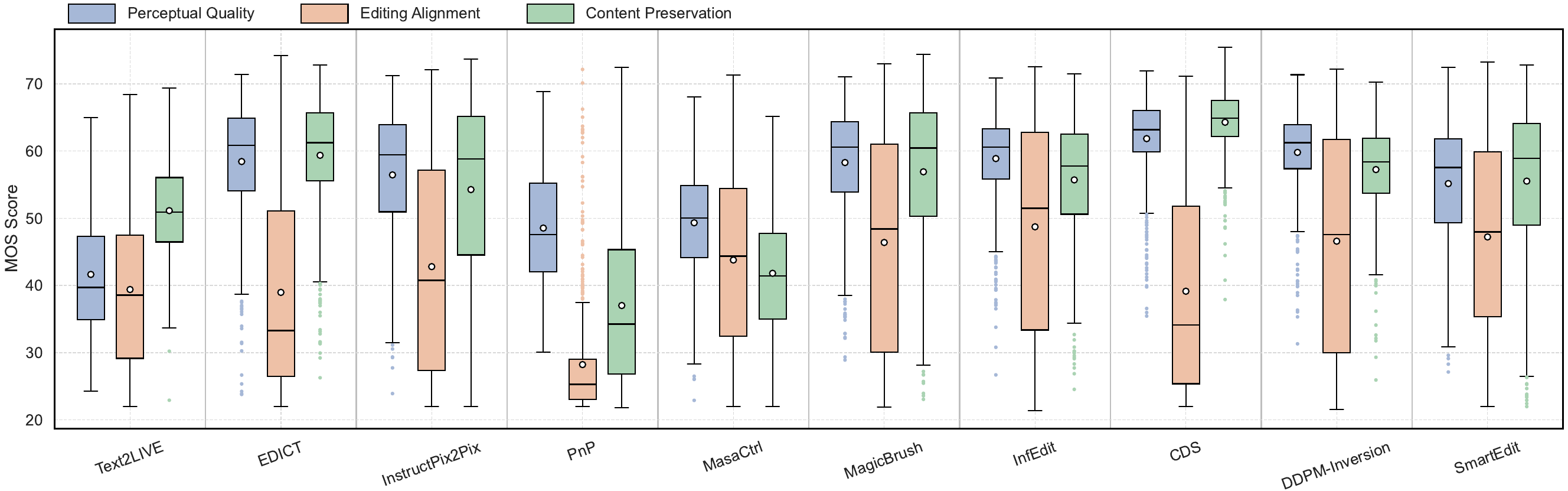}
    \vspace{-0pt}
    \caption{
MOS distributions across different text-guided image editing models.
The boxplots illustrate the performance variations of ten representative editing models across the three evaluation dimensions.
Clear performance differences can be observed among models, reflecting the diversity of editing capabilities across different editing paradigms.
}
    \vspace{-0pt}
    \label{fig:modelwise}
\end{figure*}

\textbf{Model-wise Comparison.}
We further compare the performance of different editing models using MOS distributions across models. 
As shown in Fig.~\ref{fig:modelwise}, noticeable performance differences can be observed among the ten evaluated editing models. 
Recent editing methods such as MagicBrush, InfEdit, and SmartEdit generally achieve higher median MOS scores across dimensions, indicating stronger overall editing capability. 
In contrast, earlier editing approaches such as Text2LIVE and some inversion-based editing methods exhibit larger performance variance across tasks, reflecting less stable behavior under diverse editing scenarios. 
Notably, CDS consistently achieves high content preservation scores but relatively lower editing alignment scores. 
This suggests that CDS tends to produce edits with limited deviation from the original images. 
While such behavior favors content preservation, it may restrict the model’s ability to perform substantial instruction-driven modifications, indicating a potential trade-off between content preservation and editing fidelity.

\begin{figure*}[t]
    \centering
    \includegraphics[width=0.99\linewidth]{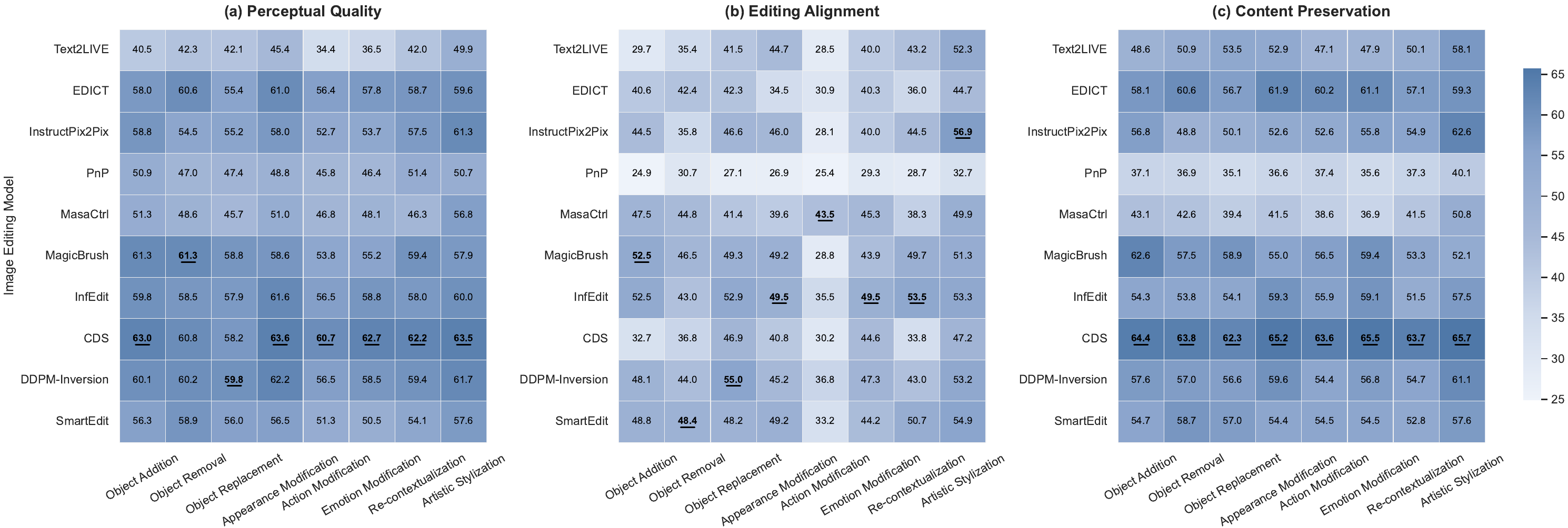}
    \vspace{-0pt}
    \caption{
Dimension-wise MOS comparison across editing models and tasks.
Each heatmap shows the average MOS values for perceptual quality, editing alignment, and content preservation across different model--task pairs.
The results reveal that editing alignment varies significantly across models and tasks, whereas perceptual quality and content preservation remain relatively stable.
}
    \vspace{-0pt}
    \label{fig:dimensionwise}
\end{figure*}

\textbf{Dimension-wise Analysis.}
Finally, we examine the interaction between editing models and editing tasks across the three evaluation dimensions. 
Fig.~\ref{fig:dimensionwise} illustrates the average MOS values for each model--task pair. 
As shown in the heatmaps, perceptual quality and content preservation remain relatively stable across models and tasks, whereas editing alignment varies significantly depending on both the model and the editing scenario. 
This result suggests that correctly interpreting and executing editing instructions remains the primary bottleneck for current text-guided image editing systems, which is consistent with observations reported in recent editing benchmarks \cite{editval,magicbrush}.

\begin{figure*}[!t]
\centering
\includegraphics[width=0.98\linewidth]{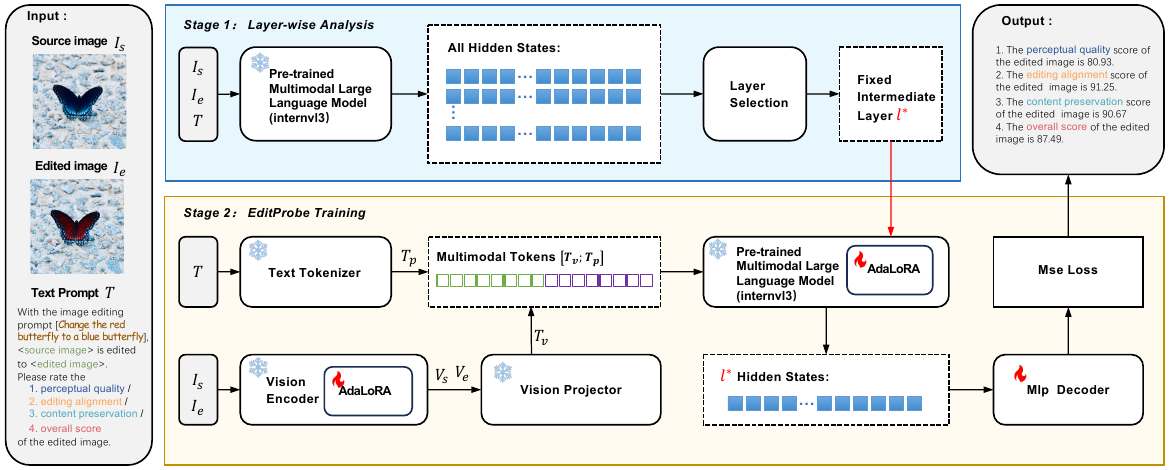}
\caption{
Overview of EditProbe.
The pipeline consists of two stages.
In Stage 1, layer-wise analysis is performed on the pre-trained MLLM to identify the most informative transformer layer for editing evaluation.
In Stage 2, the selected layer is used for intermediate-layer probing, and a lightweight regression head is trained to predict MOS scores.}
\label{fig:editprobe}
\end{figure*}

\section{EditProbe}
\label{sec:EditProbe}

We propose \textbf{EditProbe}, a probing-based evaluator for text-guided image editing built upon a pre-trained multimodal large language model (MLLM). Given a source image $I_s$, an edited image $I_e$, and an editing instruction $T$, EditProbe predicts a scalar score aligned with the corresponding human mean opinion score (MOS). Perceptual quality, editing alignment, content preservation, and overall score are modeled as four independent regression tasks and trained separately.

EditProbe consists of two stages. In \textbf{Stage 1}, we perform offline layer-wise analysis on the frozen backbone to identify the most informative transformer layer for editing evaluation. In \textbf{Stage 2}, we insert AdaLoRA modules into the backbone, extract hidden states only from the selected layer, and train a lightweight regressor for score prediction. The overall pipeline is shown in Fig.~\ref{fig:editprobe}.

\subsection{Overview}

For each sample, the source image, edited image, and editing instruction are jointly processed by the MLLM. Let
\begin{equation}
x = \{I_s, I_e, T\}.
\end{equation}

The two images are encoded by a vision encoder $E_v$:
\begin{equation}
V_s = E_v(I_s), \qquad V_e = E_v(I_e),
\end{equation}
where $V_s \in \mathbb{R}^{N_s \times d_v}$ and $V_e \in \mathbb{R}^{N_e \times d_v}$ are visual features.

The visual features are projected into the language space by a projector $P_v$:
\begin{equation}
T_v = P_v([V_s ; V_e]),
\end{equation}
where $[\,\cdot\,;\,\cdot\,]$ denotes token concatenation and $T_v \in \mathbb{R}^{N_v \times d}$.

The text instruction is tokenized as
\begin{equation}
T_p = \mathrm{Tokenizer}(T), \qquad T_p \in \mathbb{R}^{N_t \times d}.
\end{equation}

The multimodal input sequence is then formed as
\begin{equation}
Z_0 = [T_v ; T_p] \in \mathbb{R}^{(N_v + N_t)\times d}.
\end{equation}

Let $F_{\Theta}$ denote the MLLM backbone with parameters $\Theta$. A forward pass produces hidden states for all transformer layers:
\begin{equation}
H^{(1)}(x), H^{(2)}(x), \dots, H^{(L)}(x) = F_{\Theta}(Z_0),
\end{equation}
where $L$ is the number of transformer layers and
\begin{equation}
H^{(l)}(x) = [h^{(l)}_1, h^{(l)}_2, \dots, h^{(l)}_N], \qquad h^{(l)}_i \in \mathbb{R}^{d}.
\end{equation}

Instead of using the final-layer representation, EditProbe probes an intermediate layer selected by offline analysis.

\begin{figure}[!t]
    \centering
    \includegraphics[width=0.99\linewidth]{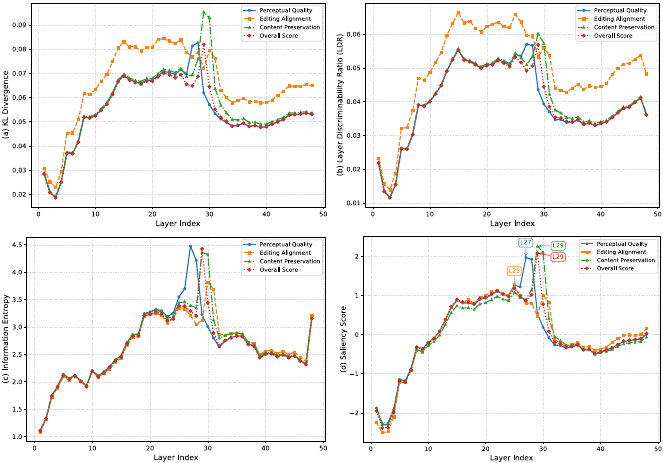}
    \vspace{-10pt}
    \caption{Layer-wise analysis of hidden state properties in the pre-trained multimodal large language model (InternVL3) on the TIEdit benchmark. KL divergence, layer discriminability ratio (LDR), information entropy, and saliency score exhibit consistent trends, with their values peaking in the intermediate transformer layers, indicating that these layers capture the most informative representations for text-guided image editing evaluation.
    }
    
    \vspace{-10pt}
    \label{fig:layer_analysis}
\end{figure}

\subsection{Stage 1: Layer Selection}
\label{sec:stage1}

Previous studies have shown that intermediate transformer layers often preserve richer semantic and structural information than the final layer \cite{raghu2017svcca, geva2021transformer}. 
Motivated by this observation, EditProbe identifies the most informative probing layer through a layer-wise representation analysis.

Stage 1 is performed once on the frozen backbone. For each training sample, we extract hidden states from \emph{all} transformer layers and evaluate the representation quality of each layer.

For layer $l$, we compute three statistics: KL divergence $KL(l)$, layer discriminability ratio $LDR(l)$, and information entropy $H(l)$. These quantities are normalized and combined into a saliency score:
\begin{equation}
S(l) = \alpha \, \widetilde{KL}(l) + \beta \, \widetilde{LDR}(l) + \gamma \, \widetilde{H}(l),
\end{equation}
where $\widetilde{KL}(l)$, $\widetilde{LDR}(l)$, and $\widetilde{H}(l)$ denote the normalized statistics, and $\alpha$, $\beta$, and $\gamma$ are weighting coefficients.
The probing layer is selected as
\begin{equation}
l^* = \arg\max_{l \in \{1,\dots,L\}} S(l).
\end{equation}

This stage uses hidden states from all layers only for offline analysis. After $l^*$ is determined, the selected layer is fixed for all subsequent training and inference.
The detailed layer-wise statistical analysis is presented in Section \ref{sec:layer-wise}.

\subsection{Stage 2: Intermediate-Layer Probing and Score Prediction}

In Stage 2, only the hidden states from the selected layer $l^*$ are used. Let $h_s^{(l^*)}$ and $h_e^{(l^*)}$ denote the hidden states corresponding to the final tokens of the source-image and edited-image visual streams after multimodal fusion. We construct the sample-level feature by averaging them:
\begin{equation}
z(x) = \frac{1}{2}\left(h_s^{(l^*)} + h_e^{(l^*)}\right),
\end{equation}
where $z(x) \in \mathbb{R}^{d}$.

This feature is fed into a lightweight MLP regressor to predict the target score:
\begin{equation}
\hat{y} = f_{\theta}(z(x)).
\end{equation}

The regression head is a two-hidden-layer MLP with ReLU activations.
Given a training set
$\mathcal{D} = \{(x_i, y_i)\}_{i=1}^{N_s}$,
where $y_i$ is the MOS label of the $i$-th sample, we minimize the mean squared error:
\begin{equation}
\mathcal{L}_{\mathrm{MSE}} = \frac{1}{B}\sum_{i=1}^{B}(\hat{y}_i - y_i)^2,
\end{equation}
where $B$ is the batch size.

\subsection{AdaLoRA Adaptation}

To adapt the backbone efficiently, we employ Adaptive Low-Rank Adaptation (AdaLoRA) \cite{zhang2023adalora}, an adaptive extension of LoRA \cite{hu2022lora}. AdaLoRA modules are inserted into both the vision encoder and the language model. The original backbone parameters remain frozen.

During training, each sample is processed by the AdaLoRA-augmented MLLM in a single forward pass. Hidden states are extracted only from the selected layer $l^*$ to form $z(x)$, and the regression loss updates only the AdaLoRA parameters and the regression head. This design preserves the pretrained multimodal prior while enabling efficient task adaptation.

\section{Experiments}
We evaluate EditProbe on the TIEdit benchmark and measure its agreement with human perceptual judgments. We first describe the experimental setting, then compare EditProbe with representative automatic evaluation metrics, followed by layer-wise analysis and ablation studies.

\subsection{Experimental Setup}

\textbf{Benchmark and protocol.}
All experiments are conducted on the TIEdit benchmark. Each edited image is annotated with three MOS dimensions: perceptual quality ($MOS_q$), editing alignment ($MOS_e$), and content preservation ($MOS_p$). The overall score ($MOS_o$) is defined as the average of the three scores.

Following standard practice in image quality assessment \cite{sheikh2006statistical}, we report Spearman rank correlation coefficient (SRCC), Pearson linear correlation coefficient (PLCC), and Kendall rank correlation coefficient (KRCC). Higher values indicate better agreement with human MOS.

We compare EditProbe with representative evaluation approaches, including full-reference IQA metrics, no-reference quality predictors, vision-language metrics, deep IQA models, and a large language model-based evaluator (GPT-4).

\begin{table*}[t]
\centering
\caption{Performance comparison of automatic evaluation methods on the \textbf{TIEdit} benchmark from four perspectives: perceptual quality, editing alignment, content preservation, and overall score. 
SRCC, PLCC, KRCC are reported to measure the correlation between metric predictions and human MOSs, where higher values indicate better agreement with human perceptual judgments. 
$\spadesuit$ denotes traditional full-reference (FR) IQA metrics, $\heartsuit$ denotes traditional no-reference (NR) IQA metrics, $\bigstar$ denotes vision--language and preference-based metrics, $\diamondsuit$ denotes deep learning-based IQA models, and $\blacktriangle$ denotes LLM-based evaluators. 
The best results in each column are highlighted in \textbf{bold}, and the second-best results are \underline{underlined}. 
The last row reports the relative improvement of EditProbe over the second-best result in each column.}
\label{tab:metric_comparison}
\resizebox{\linewidth}{!}{
\begin{tabular}{l||ccc|ccc|ccc|ccc}
\toprule
\multirow{2}{*}{\makecell[l]{Methods/Metrics}}
& \multicolumn{3}{c|}{Perceptual Quality ($MOS_q$)}
& \multicolumn{3}{c|}{Editing Alignment ($MOS_e$)}
& \multicolumn{3}{c|}{Content Preservation ($MOS_p$)}
& \multicolumn{3}{c}{Overall Score ($MOS_o$)} \\
\cmidrule(lr){2-4} \cmidrule(lr){5-7} \cmidrule(lr){8-10} \cmidrule(lr){11-13}
& SRCC & PLCC & KRCC & SRCC & PLCC & KRCC & SRCC & PLCC & KRCC & SRCC & PLCC & KRCC \\
\midrule

$\spadesuit$ FSIM~\cite{zhang2011fsim}      & 0.3906 & 0.3514 & 0.2620 & 0.2061 & 0.1419 & 0.0776 & 0.0134 & 0.0509 & 0.1152 & 0.1794 & 0.2437 & 0.3079 \\
$\spadesuit$ FSIMc~\cite{zhang2011fsim}      & 0.3896 & 0.3495 & 0.2617 & 0.2056 & 0.1417 & 0.0777 & 0.0137 & 0.0503 & 0.1142 & 0.1782 & 0.2422 & 0.3062 \\
$\spadesuit$ IFC~\cite{sheikh2005ifc}       & 0.1624 & 0.3137 & 0.1163 & 0.1514 & 0.1283 & 0.1053 & 0.0822 & 0.0592 & 0.0361 & 0.0131 & 0.0100 & 0.0330 \\
$\spadesuit$ IW-MSE~\cite{wang2011iwssim}    & 0.3976 & 0.2298 & 0.2666 & 0.0138 & 0.0517 & 0.1172 & 0.1827 & 0.2482 & 0.3137 & 0.3792 & 0.4446 & 0.5101 \\
$\spadesuit$ IW-PSNR~\cite{wang2011iwssim}   & 0.3976 & 0.3751 & 0.2666 & 0.2155 & 0.1500 & 0.0845 & 0.0190 & 0.0465 & 0.1120 & 0.1775 & 0.2430 & 0.3085 \\
$\spadesuit$ MSE~\cite{gonzalez2002digital}       & 0.3808 & 0.3698 & 0.2541 & 0.0383 & 0.1017 & 0.1650 & 0.2283 & 0.2917 & 0.3550 & 0.4184 & 0.4817 & \underline{0.5451} \\
$\spadesuit$ MS-SSIM~\cite{wang2003msssim}   & 0.4274 & 0.4018 & 0.2863 & 0.2307 & 0.1601 & 0.0895 & 0.0189 & 0.0517 & 0.1223 & 0.1929 & 0.2635 & 0.3341 \\
$\spadesuit$ NQM~\cite{damera1996nqm}       & 0.4105 & 0.4064 & 0.2752 & 0.2288 & 0.1611 & 0.0935 & 0.0258 & 0.0418 & 0.1095 & 0.1771 & 0.2447 & 0.3124 \\
$\spadesuit$ PAMSE~\cite{zeng2014pamse}     & 0.4198 & 0.4106 & 0.2822 & 0.0405 & 0.1093 & 0.1782 & 0.2470 & 0.3158 & \underline{0.3846} & \underline{0.4535} & \underline{0.5223} & 0.5911 \\
$\spadesuit$ PSNR~\cite{gonzalez2002digital}      & 0.3808 & 0.3689 & 0.2541 & 0.2079 & 0.1446 & 0.0812 & 0.0179 & 0.0454 & 0.1088 & 0.1721 & 0.2355 & 0.2988 \\
$\spadesuit$ SSIM-1~\cite{wang2004ssim}    & 0.2228 & 0.1939 & 0.1473 & 0.1125 & 0.0747 & 0.0370 & 0.0007 & 0.0385 & 0.0762 & 0.1140 & 0.1517 & 0.1895 \\
$\spadesuit$ SSIM-2~\cite{wang2004ssim}    & 0.3407 & 0.3089 & 0.2268 & 0.1782 & 0.1213 & 0.0643 & 0.0074 & 0.0495 & 0.1065 & 0.1634 & 0.2204 & 0.2773 \\
$\spadesuit$ VIF~\cite{sheikh2006vif}       & 0.1768 & 0.0761 & 0.1245 & 0.0735 & 0.0473 & 0.0212 & 0.0050 & 0.0311 & 0.0573 & 0.0835 & 0.1096 & 0.1358 \\
$\spadesuit$ VSI~\cite{zhang2014vsi}       & 0.4070 & 0.3793 & 0.2732 & 0.2194 & 0.1525 & 0.0856 & 0.0187 & 0.0482 & 0.1151 & 0.1820 & 0.2489 & 0.3158 \\
\midrule

$\heartsuit$ BMPRI~\cite{quality:BMPRI}     & 0.0307 & 0.0139 & 0.0207 & 0.0330 & 0.0433 & 0.0233 & 0.0307 & 0.0139 & 0.0207 & 0.0319 & 0.0028 & 0.0216 \\
$\heartsuit$ BPRI~\cite{min2017blind}      & 0.0005 & 0.0141 & 0.0015 & 0.0207 & 0.0449 & 0.0163 & 0.0005 & 0.0141 & 0.0015 & 0.0042 & 0.0224 & 0.0017 \\
$\heartsuit$ BRISQUE~\cite{mittal2012no}   & 0.1329 & 0.1018 & 0.0893 & 0.0890 & 0.1243 & 0.0585 & 0.1329 & 0.1018 & 0.0893 & 0.0603 & 0.0023 & 0.0394 \\
$\heartsuit$ NIQE~\cite{mittal2012niqe}      & 0.0693 & 0.0899 & 0.0478 & 0.0647 & 0.0823 & 0.0436 & 0.0694 & 0.0901 & 0.0478 & 0.1093 & 0.1311 & 0.0733 \\
$\heartsuit$ QAC~\cite{xue2013learning}       & 0.0559 & 0.0646 & 0.0358 & 0.1187 & 0.1320 & 0.0792 & 0.0559 & 0.0646 & 0.0358 & 0.0319 & 0.0461 & 0.0246 \\
$\heartsuit$ HIGRADE-1~\cite{kundu2018higrade} & 0.1253 & 0.1212 & 0.0826 & 0.0131 & 0.0614 & 0.0077 & 0.0943 & 0.1042 & 0.0635 & 0.0742 & 0.0547 & 0.0489 \\
$\heartsuit$ HIGRADE-2~\cite{kundu2018higrade} & 0.0751 & 0.0776 & 0.0495 & 0.0033 & 0.0375 & 0.0018 & 0.0919 & 0.1124 & 0.0612 & 0.0580 & 0.0447 & 0.0387 \\
\midrule

$\bigstar$ CLIPScore~\cite{clipscore}       & 0.0053 & 0.0222 & 0.0040 & 0.1357 & 0.1296 & 0.0915 & 0.0053 & 0.0222 & 0.0040 & 0.0518 & 0.0783 & 0.0331 \\
$\bigstar$ BLIPScore~\cite{li2022blip}       & 0.0200 & 0.0767 & 0.0099 & 0.1721 & 0.1681 & 0.1153 & 0.0200 & 0.0767 & 0.0099 & 0.1344 & 0.1906 & 0.0914 \\
$\bigstar$ AestheticScore~\cite{ava}  & 0.1576 & 0.1864 & 0.1057 & 0.0887 & 0.0990 & 0.0599 & 0.1576 & 0.1864 & 0.1057 & 0.1289 & 0.1456 & 0.0864 \\
$\bigstar$ ImageReward~\cite{database/align:ImageReward}     & 0.0453 & 0.0711 & 0.0300 & 0.2833 & 0.2861 & 0.1919 & 0.0453 & 0.0711 & 0.0300 & 0.1957 & 0.2316 & 0.1334 \\
\midrule

$\diamondsuit$ MANIQA~\cite{yang2022maniqa}    & 0.4755 & 0.5022 & 0.3326 & 0.3432 & 0.3440 & 0.2350 & 0.4471 & 0.4664 & 0.3164 & 0.3616 & 0.3766 & 0.2523 \\
$\diamondsuit$ CLIPIQA~\cite{zhang2023blind}   & 0.4199 & 0.4375 & 0.2604 & 0.3603 & 0.1520 & 0.2466 & 0.3858 & 0.1999 & 0.2612 & 0.3656 & 0.1540 & 0.2495 \\
$\diamondsuit$ CNNIQA~\cite{kang2014convolutional}    & 0.2729 & 0.3012 & 0.1845 & 0.2729 & 0.3012 & 0.1845 & 0.3204 & 0.3118 & 0.2157 & 0.2024 & 0.2097 & 0.1339 \\
$\diamondsuit$ DBCNN~\cite{quality:DBCNN}     & 0.4546 & \underline{0.5173} & 0.3166 & 0.3257 & 0.3287 & 0.2119 & 0.3645 & 0.4302 & 0.2477 & 0.3390 & 0.3629 & 0.2313 \\
$\diamondsuit$ HyperIQA~\cite{quality:HyperNet}  & 0.3664 & 0.3928 & 0.2497 & 0.2813 & 0.3106 & 0.1932 & 0.2946 & 0.3224 & 0.1944 & 0.2946 & 0.3258 & 0.1987 \\
$\diamondsuit$ NIMA~\cite{talebi2018nima}      & 0.2412 & 0.2487 & 0.1617 & 0.2171 & 0.2087 & 0.1462 & 0.1591 & 0.1441 & 0.1064 & 0.1683 & 0.1537 & 0.1145 \\
$\diamondsuit$ TOPIQ~\cite{topiq}     & 0.4473 & 0.4844 & 0.3117 & 0.3638 & 0.3335 & 0.2487 & 0.3536 & 0.4053 & 0.2394 & 0.4078 & 0.4087 & 0.2832 \\
$\diamondsuit$ WaDIQaM~\cite{bosse2017deep}   & 0.2763 & 0.2871 & 0.1876 & 0.1188 & 0.1101 & 0.0797 & 0.3291 & 0.3286 & 0.2212 & 0.2558 & 0.2665 & 0.1734 \\
$\diamondsuit$ VGG16~\cite{simonyan2014very}     & \underline{0.5323} & 0.5057 & \underline{0.3683} & 0.4726 & 0.4439 & 0.3310 & \underline{0.5955} & \underline{0.5228} & \underline{0.4181} & 0.4944 & 0.4850 & 0.3395 \\
$\diamondsuit$ VGG19~\cite{simonyan2014very}     & 0.5268 & 0.4805 & 0.3623 & \underline{0.4837} & \underline{0.4549} & \underline{0.3392} & 0.5829 & 0.5198 & 0.4122 & \underline{0.4982} & 0.4777 & 0.3340 \\
\midrule

$\blacktriangle$ GPT-4~\cite{openai2023gpt4}            & 0.0084 & 0.0065 & 0.0653 & 0.2230 & 0.1664 & 0.2013 & 0.0779 & 0.0582 & 0.0875 & 0.0592 & 0.0413 & 0.1638 \\
\midrule

\rowcolor{gray!15}
\textbf{EditProbe}
& \textbf{0.6959} & \textbf{0.7064} & \textbf{0.5095}
& \textbf{0.7995} & \textbf{0.7910} & \textbf{0.5974}
& \textbf{0.7715} & \textbf{0.7892} & \textbf{0.5887}
& \textbf{0.7795} & \textbf{0.7807} & \textbf{0.5913} \\

\rowcolor{gray!10}
Improvement
& +30.7\% & +36.6\% & +38.3\%
& +65.3\% & +73.9\% & +76.1\%
& +29.6\% & +50.9\% & +40.8\%
& +56.5\% & +49.5\% & +8.5\% \\
\bottomrule
\end{tabular}}
\end{table*}

\textbf{Implementation details.}
EditProbe uses InternVL3 \cite{zhu2025internvl3} as the backbone MLLM. The source image, edited image, and editing instruction are processed jointly. Hidden states from the selected transformer layer are used as regression features. The regression head is a two-hidden-layer MLP trained with MSE loss.

\textbf{Training settings.}
We adopt AdaLoRA \cite{zhang2023adalora} for parameter-efficient adaptation of both the vision encoder and the language model. The initial rank is 16, the scaling factor is 32, and the dropout rate is 0.05. Only the AdaLoRA parameters and regression head are updated during training, while the backbone remains frozen.
All images are resized to $448\times448$ and normalized with ImageNet statistics. Optimization uses AdamW with an initial learning rate of $1\times10^{-5}$ and cosine decay with warm-up. The batch size is 8 and the model is trained for 5 epochs on a 40GB NVIDIA RTX A100 GPU.

\subsection{Performance Comparison}
Table~\ref{tab:metric_comparison} reports the comparison between EditProbe and existing evaluation metrics.
Traditional full-reference IQA metrics such as PSNR and SSIM measure pixel-level similarity and therefore cannot properly evaluate semantic edits required by text instructions. Vision--language metrics partially capture instruction compliance but do not explicitly model the relationship between the source image and the edited result. Deep IQA models achieve better performance but are typically trained on distortion datasets rather than editing tasks.

EditProbe achieves the highest correlation with human MOS on all four evaluation dimensions. Specifically, it obtains SRCC values of 0.6959, 0.7995, 0.7715, and 0.7795 for perceptual quality, editing alignment, content preservation, and overall score, respectively. The improvement over the strongest baselines demonstrates the effectiveness of intermediate-layer probing for editing evaluation.

We also observe that directly using a general-purpose MLLM evaluator is insufficient. GPT-4 shows substantially lower correlation with human MOS, suggesting that reliable editing evaluation depends on task-specific representations rather than free-form multimodal reasoning alone.

\subsection{Layer-wise Representation Analysis}
\label{sec:layer-wise}

To validate the layer selection strategy in Section \ref{sec:stage1} of Stage 1, we analyze hidden states across transformer layers using KL divergence, layer discriminability ratio (LDR), information entropy, and the saliency score.

Figure~\ref{fig:layer_analysis} shows a consistent pattern across all four indicators. KL divergence and LDR increase from shallow layers and peak in the intermediate layers, indicating stronger separability between samples with different editing quality levels. Entropy follows the same trend, suggesting that intermediate layers preserve richer representations. The saliency score reaches its maximum in the same region. In deeper layers, all indicators decline slightly. This indicates that later layers become increasingly specialized for autoregressive generation and are less suitable for evaluation. These observations justify the use of intermediate-layer probing instead of the final-layer representation.

\begin{table*}[ht]
\centering
\caption{
Ablation study of different design choices in \textbf{EditProbe} on the \textbf{TIEdit} benchmark from four perspectives: perceptual quality, editing alignment, content preservation, and overall score.
SRCC, PLCC, and KRCC measure the correlation between metric predictions and human MOSs, where higher values indicate better agreement.
The best results in each column are highlighted in \textbf{bold}, and the second-best results are \underline{underlined}.
}
\label{tab:ablation}
\resizebox{\linewidth}{!}{
\begin{tabular}{l||ccc|ccc|ccc|ccc}
\toprule
\multirow{2}{*}{\makecell[l]{Methods}}
& \multicolumn{3}{c|}{Perceptual Quality ($MOS_q$)}
& \multicolumn{3}{c|}{Editing Alignment ($MOS_e$)}
& \multicolumn{3}{c|}{Content Preservation ($MOS_p$)}
& \multicolumn{3}{c}{Overall Score ($MOS_o$)} \\
\cmidrule(lr){2-4} \cmidrule(lr){5-7} \cmidrule(lr){8-10} \cmidrule(lr){11-13}
& SRCC & PLCC & KRCC & SRCC & PLCC & KRCC & SRCC & PLCC & KRCC & SRCC & PLCC & KRCC \\
\midrule

w/o MLP
& 0.6594 & 0.6644 & 0.4764
& 0.7624 & 0.7464 & 0.5588
& 0.7205 & 0.7522 & 0.5374
& 0.7296 & 0.7335 & 0.5407 \\

LoRA (only LLM)
& 0.6775 & 0.6876 & 0.4932
& 0.7814 & 0.7712 & 0.5778
& 0.7456 & 0.7673 & 0.5630
& 0.7567 & 0.7567 & 0.5671 \\

LoRA (only Vision)
& 0.6899 & 0.6994 & 0.5039
& 0.7926 & 0.7833 & 0.5908
& 0.7656 & 0.7826 & 0.5822
& 0.7708 & 0.7737 & 0.5833 \\

Final Layer
& \underline{0.6953} & \underline{0.7056} & \underline{0.5087}
& \underline{0.7967} & \underline{0.7876} & \underline{0.5939}
& \underline{0.7710} & \underline{0.7888} & \underline{0.5876}
& \underline{0.7771} & \underline{0.7794} & \underline{0.5897} \\

\rowcolor{gray!15}
\textbf{EditProbe}
& \textbf{0.6959} & \textbf{0.7064} & \textbf{0.5095}
& \textbf{0.7995} & \textbf{0.7910} & \textbf{0.5974}
& \textbf{0.7715} & \textbf{0.7892} & \textbf{0.5887}
& \textbf{0.7795} & \textbf{0.7807} & \textbf{0.5913} \\

\bottomrule
\end{tabular}}
\end{table*}

\subsection{Ablation Study}
Table~\ref{tab:ablation} summarizes the ablation results. Removing the regression head (w/o MLP) reduces performance on all dimensions, showing that a learned regressor is necessary to map hidden representations to MOS. Applying AdaLoRA only to the language model or only to the vision encoder also leads to performance degradation, which indicates that editing evaluation depends on both visual modeling and cross-modal reasoning.

We further compare the selected intermediate layer with the final transformer layer. Using the final layer already yields reasonable performance, but the selected intermediate layer is consistently better. For overall score prediction, SRCC improves from 0.7296 to 0.7795. This confirms that intermediate representations are more informative than final-layer representations for editing evaluation.

\section{Conclusion}

This paper presents TIEdit, a benchmark for evaluating text-guided image editing with large-scale human annotations across perceptual quality, editing alignment, and content preservation. Using this benchmark, we conduct a systematic study of existing automatic evaluation metrics and show that their correlations with human judgments are limited for editing tasks.
To address this gap, we introduce EditProbe, an evaluation framework that leverages intermediate representations of multimodal large language models to predict editing quality. Our analysis reveals that editing-related information is most discriminative in intermediate transformer layers, motivating the probing strategy adopted in EditProbe.
Extensive experiments demonstrate that EditProbe achieves substantially stronger agreement with human perceptual judgments than existing evaluation methods. We hope that TIEdit and EditProbe provide a foundation for more reliable evaluation of text-guided image editing and enable future progress in multimodal generative modeling.

\bibliographystyle{IEEEtran} 

\bibliography{arxiv}



\end{document}